\documentclass[a4paper,conference]{IEEEtran}
\usepackage{cite}
\usepackage{svg}
\usepackage{amsmath,amssymb,amsfonts}
\usepackage{algorithmic}
\usepackage{graphicx}
\usepackage{textcomp}
\usepackage{xcolor}
\usepackage{times}
\usepackage{graphicx}
\usepackage{amsmath}
\usepackage{amssymb}
\graphicspath{ {img/} }
\usepackage{hyperref}
\usepackage{subcaption}
\usepackage{float}
\usepackage{xcolor}
\usepackage{url}
\usepackage{pdfpages}
\usepackage{fancyhdr}
\IEEEoverridecommandlockouts

\fancypagestyle{plain}{
    \fancyhf{}
    \fancyhead[L]{\textcolor{red}{M. Dale, A. Ross, A. Jain, “On Missing Scores in Evolving Multibiometric Systems,” Proc. of 26th International Conference on Pattern Recognition (ICPR), (Montreal, Canada), August 2022.}}
}

\newcommand{\biocop}{BIOCOP2008}

\begin{document}
\title{On Missing Scores in Evolving Multibiometric Systems}

\author{\IEEEauthorblockN{Melissa R Dale}
\IEEEauthorblockA{
Michigan State University\\
East Lansing, Michigan 48824\\
Email: dalemeli@msu.edu}

\and
\IEEEauthorblockN{Anil Jain}
\IEEEauthorblockA{
Michigan State University\\
East Lansing, Michigan 48824\\
Email: jain@cse.msu.edu}

\and
\IEEEauthorblockN{Arun Ross}
\IEEEauthorblockA{Michigan State University\\
East Lansing, Michigan 48824\\
Email: rossarun@msu.edu}
}

\maketitle
\thispagestyle{plain}

\begin{abstract}
The use of multiple modalities (e.g., face and fingerprint) or multiple algorithms (e.g., three face comparators) has shown to improve the recognition accuracy of an operational biometric system. Over time a biometric system may evolve to add new modalities, retire old modalities, or be merged with other biometric systems. This can lead to scenarios where there are missing scores corresponding to the input probe set. Previous work on this topic has focused on either the verification or identification tasks, but not both. Further, the proportion of missing data considered has been less than 50\%. In this work, we study the impact of missing score data for both the verification and identification tasks. We show that the application of various score imputation methods along with simple sum fusion can improve recognition accuracy, even when the proportion of missing scores increases to 90\%. Experiments show that fusion after score imputation outperforms fusion with no imputation. Specifically, iterative imputation with K nearest neighbors consistently surpasses other imputation methods in both the verification and identification tasks, regardless of the amount of scores missing, and provides imputed values that are consistent with the ground truth complete dataset.  
\end{abstract}

\section{Introduction}
Biometrics refers to the task of recognizing individuals based on their unique physical or behavioral traits \cite{ross2006handbook}\footnote{The terminology used in this article follows current recommendations for multibiometric fusion defined by International Organization for Standardization [ISO], 2015 \cite{iso_2015}}. A typical biometric {\em identification} system aims to determine a potential set of matching identities, while a {\em verification} system aims to determine if a probe's claimed identity is valid (i.e., genuine) or not (i.e., imposter). To accomplish either goal, the feature set extracted from the probe is compared against a gallery template to produce a similarity score. 
% Biometrics refers to the task of recognizing individuals based on their unique physical or behavioral traits \cite{ross2006handbook}\footnote{\green{The terminology used in this article follows current recommendations for multibiometric fusion defined by the International Organization for Standardization \cite{iso_2015}}}. A typical biometric {\em identification} system compares the probe signal (e.g., a fingerprint image) with the gallery templates in a database (e.g., fingerprint templates, each labeled with a subject identifier) to determine a potential set of matching identities, while a {\em verification} system aims to determine if a probe's claimed identity is valid (i.e., genuine) or not (i.e., imposter). To accomplish either goal, the feature set extracted from the probe is compared against a gallery template to produce a similarity score. In verification tasks, these similarity scores are compared against a threshold to determine if the probe's claimed identity is genuine or an imposter. In identification tasks, scores are sorted to determine those gallery identities with the highest degree of similarity to the probe.  In the open set identification, where the probe may or may not be in the gallery, the highest similarity score is compared against a threshold in order to determine if there is a match or not. 

Operational or deployed biometric systems evolve as technology evolves. For example, new more advanced sensors may be added to existing systems or outdated sensors removed. In these updated systems, not all enrolled (gallery) identities will have data pertaining to all the modalities. Additionally, consider a scenario where multiple operational biometric systems are to be merged into a single biometric system. The list of available gallery identities for the different modalities may not be the same. In situations like these, the resulting gallery data is likely to be more sparse than the typical biometric galleries often studied in literature. In this paper we explore how to handle missing similarity scores of a probe through a combination of imputation \footnote{In statistics, imputation refers to the process of replacing missing values with an estimated value} and score fusion.

\section{Background}
\label{sec:Background}
\subsection{Biometric Systems}
% A typical biometric system has a sensor module, a feature extraction module, a comparator module, and a database (gallery) module.   

% Figure \ref{fig:biosys} illustrates these components.

% \begin{figure}

% \centering
% \includegraphics[width=0.5\textwidth]{bio.pdf}
% \caption{Illustration of the components present in a typical biometric system.}

% \label{fig:biosys}
% \end{figure}

Biometric systems using a single biometric modality are sometimes referred to as \textit{unimodal} systems. Relying on a single biometric cue can negatively impact biometric system performance when the input data (probe) is of low quality, such as low-resolution faces, highly distorted fingerprints, and occluded irides. Additionally, disabilities or illnesses may prevent some subjects from providing a  a biometric cue \cite{boulgouris2009biometrics, mishra2010multimodal}.  The use of multiple biometric cues in a biometric system is often referred to as \textit{multibiometrics} and can help address these issues. Leveraging multibiometric approaches has shown to improve recognition performance, enhance system reliability, combat noisy sensor data, address large variations in user samples, and improve the system's security \cite{brunelli1995person, prabhakar2002decision, toh2004exploiting, ross2003information}.  However, there are also potential drawbacks if the multiple biometric sources are not combined carefully, including an increase in recognition error, the need for additional sensors, longer recognition times, and sometimes lower user convenience \cite{allano2010tuning}. 

 Once multiple sources of biometric information are identified for an application, the critical question is how to best leverage the information from these sources. \textit{Information fusion} has been used to improve classification performance in classical machine learning and has been popular in multibiometrics \cite{kittler1998combining, ross2003information, singh2019comprehensive}. While fusion can occur at any point in the biometric system, this paper focuses on fusion at the similarity score level. Score fusion with multimodal data has been shown to increase the performance of biometric systems \cite{nandakumar2007likelihood, snelick2005large, jain2005score}. There are many existing methods to fuse similarity scores across multiple biometric modalities. One popular fusion technique is the {\em simple sum rule}, which takes the mean of available scores to produce a fused score \cite{ross2006handbook}. The simple sum fusion approach is a transformation-based approach to fusion, as the scores are first transformed into a common domain prior to fusing (e.g., scores are normalized into the range [0-1]). Simple sum fusion is a popular choice because of its straightforward approach, flexibility in the data it can be applied to, and it often produces desired results. The work in this paper utilizes the simple sum fusion approach.   

% Additional approaches to score level fusion include training a classifier such as a support vector machine (SVM) \cite{fierrez2005discriminative} or a random forest classifier \cite{ma2005classification}. However, these approaches rely upon selecting the appropriate classifier to use, as well as the associated parameter settings. Additionally, classifier-based fusion approaches can add additional complexities as the size of the biometric system grows.  Yet another approach to score fusion uses the likelihood ratio of score modalities \cite{nandakumar2007likelihood}. The performance of density-based score fusion relies heavily upon accurate estimates of score distributions and can be impacted by lack of scores. 

When performing multimodal fusion, it should be noted that auxiliary information may also be combined with score data to improve performance. This auxiliary information can include quality measures \cite{ding2016bayesian, liu2017quality},  soft biometrics \cite{kazemi2018attribute}, and demographic information \cite{sultana2017social, paul2014decision}. Incorporating additional information into the fusion can require additional design decisions and computational resources.

The papers discussed above focus on the verification task (one-to-one comparison), which is a binary classification problem (e.g., ``accept” or ``reject”). Identification tasks (one-to-many comparisons) are often more complex due to the multi-class nature of the problem. Nandakumar et al. successfully extend the  likelihood ratio fusion from verification tasks to identification tasks \cite{nandakumar2009fusion}. However, to achieve competitive performances on the identification task, the authors propose a hybrid fusion approach which not only leverages the similarity scores but also incorporates the rank information in the fusion.  The authors demonstrate the ability of this approach to handle sparse data by randomly dropping 5\%, 10\%, and 25\% of all partitions in the NIST BSSR1 dataset. 

% Challenges associated with missing score data are explored in Subsection \ref{sec:missing}. 

\subsection{Fusion with Missing Scores}
\label{sec:missing}
Missing scores can present a challenge when designing a multibiometric system since many fusion methods require score data to be complete. In deciding how to handle missing data, it is important to consider \textit{why} the data is missing. Patterns of missing values are defined by Rubin in \cite{rubin1976inference}: 

\begin{itemize}
\item \textbf{Missing Completely at Random (MCAR):} MCAR describes missing values where the probability that a value is missing is unaffected by other data, whether observed or unobserved. For example, a patient's missing weight value cannot be explained by observed data such as their sex or age, or unobserved data such as a scale's battery malfunctioning. This missing value is MCAR.

\item \textbf{Missing at Random (MAR):} Values that are MAR are influenced by observed data. Bhaskaran and Smeeth highlight MAR by providing the example of blood pressure records \cite{bhaskaran2014difference}. Records for older people are more likely to be documented because it is more often a regimen of their care. While blood pressure records for younger people may be more sparse compared to their elderly counterparts, this difference can be explained by the observable data of age.     
% Missing at random means there might be systematic differences between the missing and observed values, but these can be entirely explained by other observed variables.

\item \textbf{Missing Not at Random (MNAR):} If values are neither MCAR nor MAR, missing values are MNAR. An example of a value MNAR is if a patient's drug test is missing \textit{because} they intentionally skipped in order to prevent a positive test value. 
\end{itemize} 

Many methods of addressing missing data require the data to be either MCAR or MAR because MNAR has the potential to skew the data. If data can be assumed to be either MCAR or MAR, the following approaches may be applied.

One option for missing data is to simply ignore probes with incomplete scores, i.e., probes which do not have scores corresponding to {\em all} the modalities. This approach works if there are only a small proportion of missing data and that missing data is truly MCAR. Additionally, this approach may not be an effective use of available data if otherwise usable scores are dropped. For example in the multibiometrics context, consider a probe that obtains a similarity score for the face and iris modalities, but is missing a fingerprint similarity score. When the entire probe is removed from  consideration, a valid face similarity score and a valid iris similarity score are dropped from the dataset. Ignoring an entire probe from analysis is referred to as \textit{Listwise Deletion} \cite{kang2013prevention}. When deleting data, it is also important to consider if deleting said data would result in bias. For example, if a sensor in a biometric system is more likely to fail than other sensors in the system, missing scores are not missing at random, but rather likely missing due to the sensor used to collect the data.

Another option to handling missing data is to simply fill missing scores with the modality's mean or median score value. This univariate approach only requires information about the missing modality's scores and is unaffected by other modalities. Again consider a situation like that described above and presented in Table \ref{tab:demo}. Using mean substitution, for example, would replace the face modality's missing score with $0.51$ (the mean of the given score: $0.51$=$\frac{0.41+0.27+0.85}{3}$) and the missing fingerprint score would be replaced with $0.54$. Likewise, the missing scores may be imputed with the median. For this example the face modality's missing score is replaced with the median of the face scores $0.25$ and the missing fingerprint score would be replaced with $0.74$.

% (the median of the given score's: $Median(x)$=$\frac{Face[\frac{3-1}{2}]}{2} + Face[\frac{3+1}{2}]{2}$) and the missing Fingerprint score would be replaced with $0.74$.

\begin{table}[]
\centering
\caption{A simple example of a score dataset with missing values, denoted as ?.}
\begin{tabular}{|c|c|c|c|}
\hline 
 \textbf{Subject} & \textbf{Face} &  \textbf{Fingerprint} &  \textbf{Iris} \\\hline 

Subject 1 &  ? &         0.74 &  1.00 \\\hline 
Subject 2 & 0.41 &         0.89 &  0.47 \\\hline 
Subject 3 & 0.27 &          ? &  0.03 \\\hline 
Subject 4 & 0.85 &         0.00 &  0.31 \\\hline 

\end{tabular}
\label{tab:demo}
\end{table}

Multivariate imputation schemes attempt to map relationships between the modalities. \textit{Multiple Imputations by Chained Equations } (MICE) is a popular approach to multiple imputations, where missing values are temporarily filled with a placeholder value and then iteratively updated using a trained machine learning model \cite{van2011mice}. In the given example, shown in Table \ref{tab:demo}, both the face and iris missing values are initially filled with each modality's mean or median. The scores of individual modalities are sequentially and iteratively updated with a specified machine learning classifier. Once the classifier has been trained, the missing values are updated from the initial placeholder value to the value predicted by the trained classifier, and then the next modality's scores are fixed and the classifier is trained again to update the placeholder values. This process is repeated for a specified number of iterations, or until the imputed values stop changing between iterations.

Missing data within individual biometrics can occur in many different ways, as described in the introduction. In order to better understand the role of missing scores in multibiometric settings, researchers often are required to simulate datasets with missing score values \cite{nandakumar2009fusion, tran2011approach, ding2012comparison, poh2007biosecure, fatukasi2008estimation, damer2013missing}. In each of these prior studies, missing scores are simulated by randomly dropping scores from complete datasets. The percentage of dropped scores range from 5\% to 50\%. Simulating missing scores allows researchers to know that missing values truly are MAR, and allows for a fine degree of control over the amount of missing data. However, in real world situations, such as the example of incorporating a new modality to an existing biometric system, the proportion of missing data is likely to exceed 50\%.  In addition to the likelihood estimation approach proposed by \cite{nandakumar2007likelihood}, additional proposed multimodal imputation methods include classifier based methods such as SVM \cite{park2007iris}, or Na\"ive Bayes \cite{aravinth2016multi}, and deep learning approaches \cite{maity2020multimodal}.

In this work, we focus on Listwise Deletion, Mean (and Median) Substitution, and MICE to understand how performances may vary between simulated missing score data and real world datasets that naturally contain missing score values. We explore how these methods perform as the amount of missing score data increases in multiple biometric system evolution scenarios. 

\section{Experiments}
\label{sec:experiments}
We present 3 sets of experiments modelling 3 real-world biometric scenarios: adding a new modality to an existing multimodal biometric system, merging separate biometric systems into one, and retiring a modality from an existing multimodal system. To facilitate these experiments, we leverage the NIST BSSR1 dataset\cite{BSSR1nist}.\footnote{NIST BSSR1 dataset is available at \url{https://www.nist.gov/itl/iad/image-group/nist-biometric-scores-set-bssr1}} This publicly available multimodal dataset is comprised of similarity scores for 4 modalities: one score for the comparison of the user's right index finger to each gallery identity's right index  finger, one score for comparison of the user's left index  finger to each gallery identity's left  index finger, and two scores reported from two facial comparators (referred to as Face Algorithm C and Face Algorithm G). 

\begin{figure}
    \centering
    \includegraphics[width=0.4\textwidth]{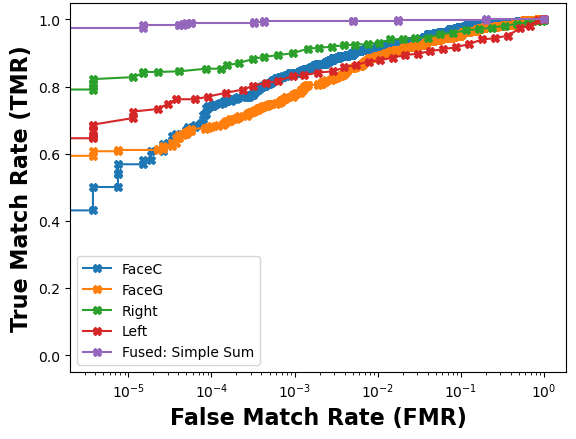}
    \caption{ROC Curves for the complete NIST BSSR1 dataset }
    \label{fig:NIST-roc}
\end{figure}

The NIST BSSR1 Set 1 dataset contains complete score vectors for each of the 517 user comparisons to the 517 templates in the gallery. Table \ref{tab:nist-scores} provides a summary of the scores in this dataset. Figure \ref{fig:NIST-roc} shows the receiver operating characteristic (ROC) curves for the complete NIST BSSR1 dataset. 

% The Rank 1 accuracies for the complete set are as follows: Right Fingerprint 93.04\%, Left Fingerprint 86.27\%, Face G Matcher 85.69\%, Face C Matcher 88.39\%, and Fused 99.81\%.

\begin{table}[]
\centering
\caption{Summary of the scores present in set1 of the NIST BSSR1 multimodal dataset.}
\begin{tabular}{|l|l|l|}
\hline
    & \textbf{Per Modality} & \textbf{Total} \\ \hline
\textbf{Total Number of Scores}     & 267,289     & 1,069,156 \\ \hline
\textbf{Genuine Scores}       & 517   & 2,068 \\ \hline
\textbf{Imposter Scores}      & 266,772   & 1,067,088        \\ \hline

% \textbf{\% Missing Scores}      & 0.0   & 0.0  \\ \hline
% \textbf{Number of Probe Identities} & 517       & 56,638          & 56,652\\ \hline
% \textbf{Score Range}   & {[}0-31,017{]} & {[}524-8,476{]} & {[}0-1,077,794{]} \\ \hline
\end{tabular}
\label{tab:nist-scores}
\end{table}

\begin{figure}[h!]
    \includegraphics[width=.3\textwidth]{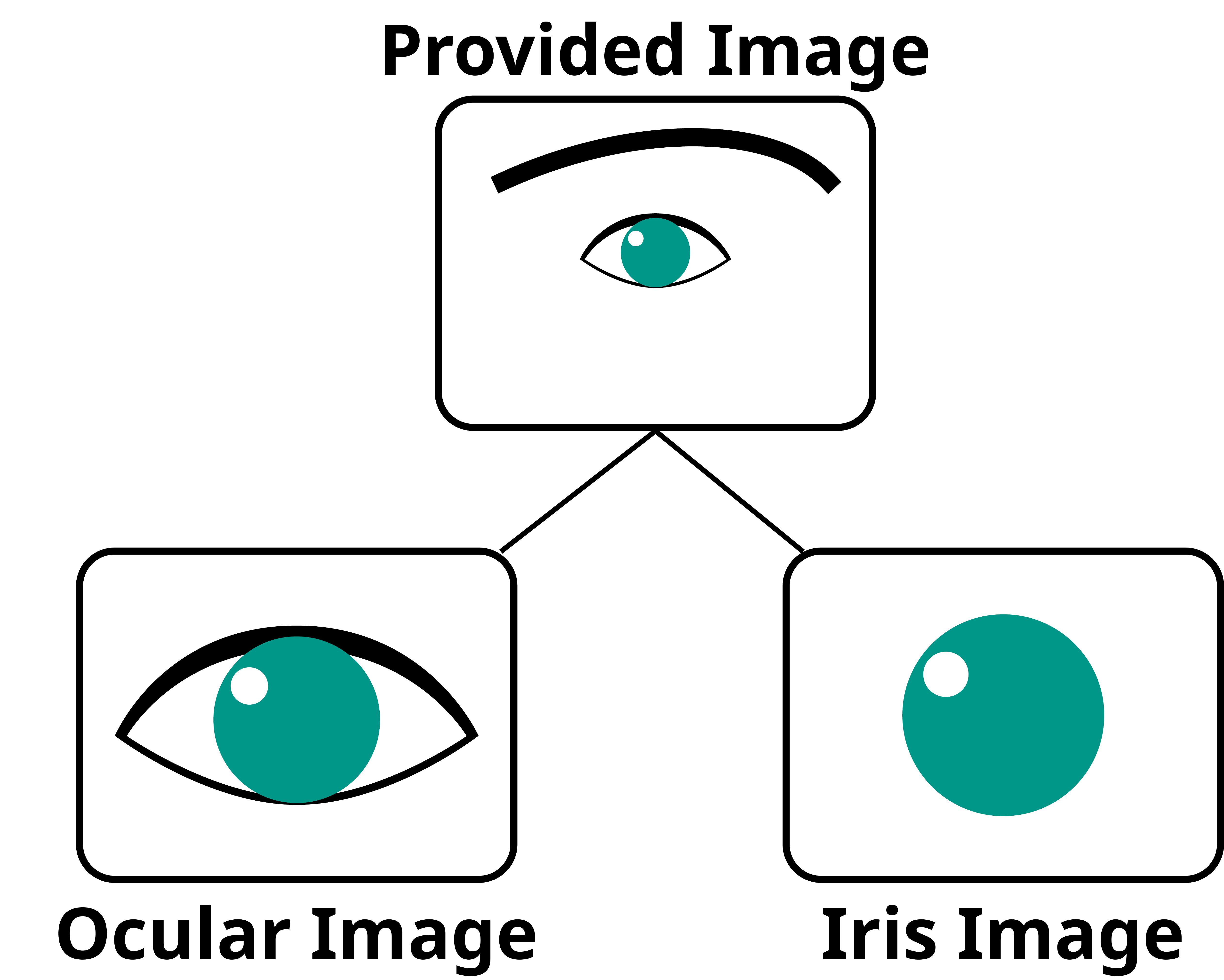}
    \centering
    \caption{Breakdown of provided Ocular image in the \biocop\ dataset}
    \label{fig:Ocular}
\end{figure}

In addition to the NIST BSSR1 dataset, we generate from the \biocop\ Ocular dataset. Similarity scores were obtained for both the ocular image and iris image using a trained Multi-Channel CNN.\footnote{Note that scores here are generated using a sub-optimal technique for the purposes of this experiment.} Figure \ref{fig:Ocular} summarizes this breakdown of the provided images.
 
\begin{figure}
    \centering
    \includegraphics[width=0.4\textwidth]{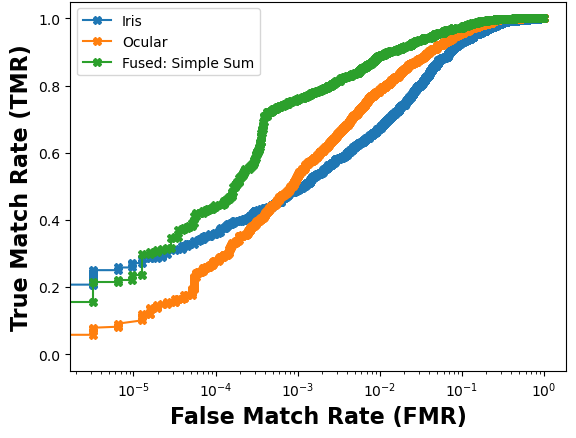}
    \caption{ROC Curves for the complete \biocop\ dataset}
    \label{fig:BIOCOP-roc}
\end{figure}

If we consider ``Ocular'' and ``Iris'' as the two modalities in this dataset, the resulting score dataset is complete. That is, for every iris image, there is a corresponding ocular image and vice versa. This is logical, as each iris image is a subset of a corresponding ocular image. This 2-modality formatting is described in Table \ref{tab:biocop2-scores}. Figure \ref{fig:BIOCOP-roc} shows the ROC curves for the complete \biocop \ dataset. 

% The Rank 1 accuracies for the Iris and Ocular modalities are 56.4\% and 66.9\%,  respectively. The simple sum fusion provides a Rank 1 identification accuracy of 78.47\%.

\begin{table}[H]
\centering
\caption{Summary of the scores present in the \biocop\ ocular dataset.}
\begin{tabular}{|l|l|l|}
\hline 
 & \textbf{Per Modality} &    \textbf{Total} \\ \hline 

 \textbf{Total Number of Scores} &    312,606 &  625,212 \\ \hline 
 \textbf{Genuine Scores} &   998 &    1,996 \\ \hline 
\textbf{ Imposter Scores} &   311,608 &  623,216 \\ \hline 

\end{tabular}
\label{tab:biocop2-scores}
\end{table}

% However if we instead break this dataset into 4 modalities (``Ocular-Left'', ``Ocular-Right'', ``Iris-Left'' and ``Iris-Right''), we find that while most users do have a match score for each of these 4 modalities, a number of users do not possess both left and right ocular images resulting in a naturally occurring missing scores within the score matrix.  

% \begin{table}[H]
% \centering
% \caption{Summary of the scores present in the \biocop\-MultiChannel 4-Modality Score Dataset.}
% \begin{tabular}{|l|l|l|l|l|}
% \hline
%     & \textbf{\begin{tabular}[c]{@{}l@{}}Iris \\ (Left)\end{tabular}} 
%     & \textbf{\begin{tabular}[c]{@{}l@{}}Iris \\ (Right)\end{tabular}} 
%     & \textbf{\begin{tabular}[c]{@{}l@{}}Ocular \\ (Left)\end{tabular}} 
%     & \textbf{\begin{tabular}[c]{@{}l@{}}Ocular \\ (Right)\end{tabular}} \\ \hline
    
% \textbf{Total Scores}     & 183,762
%      & 149,365 & 183,762 & 149,365

%  \\ \hline
% \textbf{Genuine Scores}       & 554  & 471 & 554  & 471  \\ \hline
% \textbf{Imposter Scores}      & 183,208 & 148,894  & 183,208 & 148,894  \\ \hline

% \textbf{Missing Scores}      & 6.25\%    & 23.8\%  & 6.25\%    & 23.8\%   \\ \hline
% \end{tabular}
% \label{tab:biocop2-scores}
% \end{table}

\begin{figure}

\centering
\includegraphics[width=0.4\textwidth]{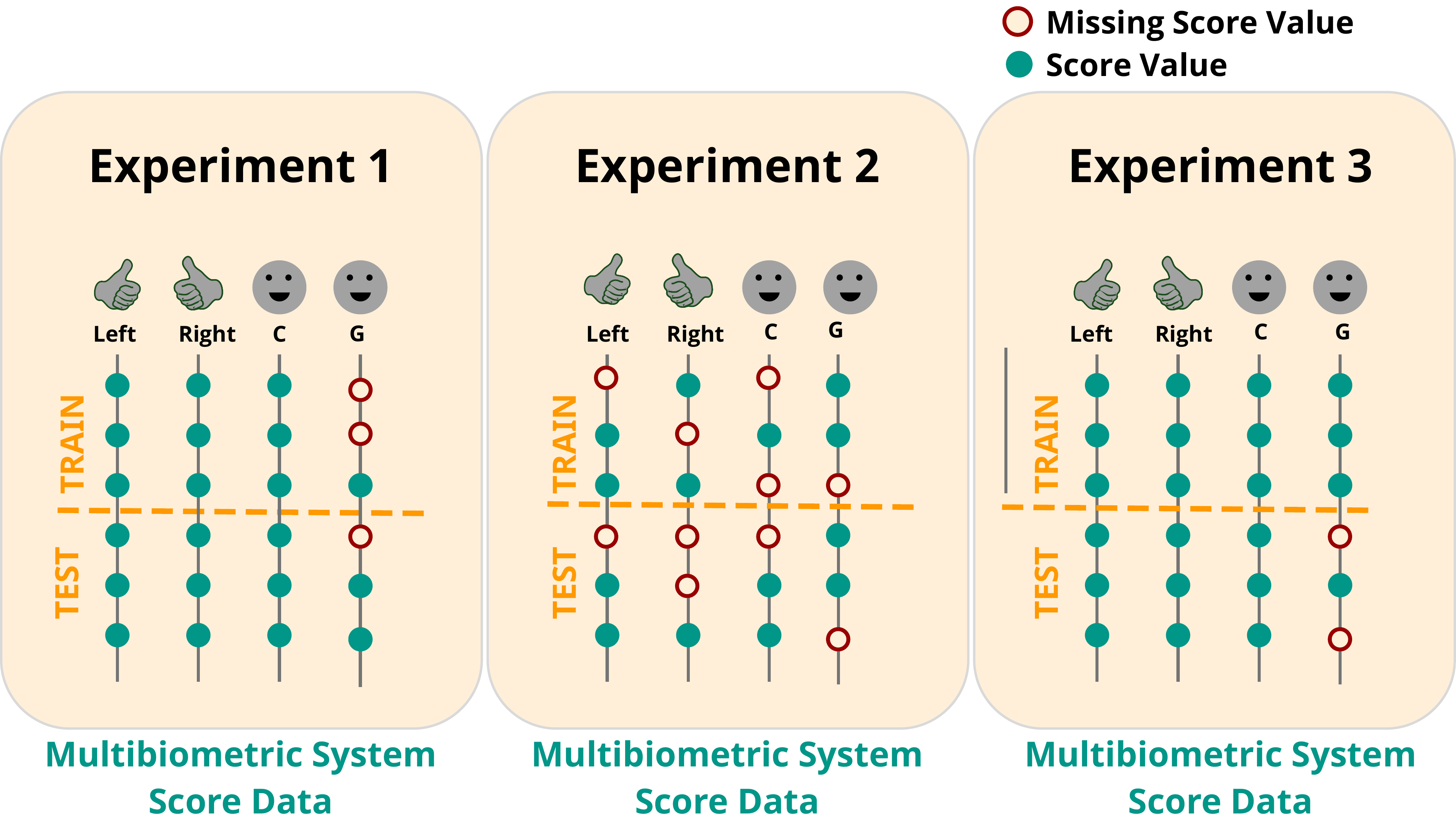}
\caption{Experiment contexts for NIST BSSR1 dataset.}

\label{fig:exps}
\end{figure}

Each experiment is described in detail below. The experimental design for all experiments is summarized in Table \ref{tab:experiments} and a diagram of the experiments is presented in Figure \ref{fig:exps}. To perform score fusion, we opt to apply the simple sum fusion as described in Section \ref{sec:Background}. This fusion technique allows us to compare the performance of fusion with imputed scores to the performance of incomplete score vectors. Many other fusion methods (e.g., score fusion with SVM \cite{park2007iris}, or Na\"ive Bayes \cite{aravinth2016multi})  a complete score vector is required and incomplete score vectors would have to be ignored (listwise deletion).  We divide the data into train and test sets, and randomly drop the specified proportion of scores 5 times to obtain mean and standard deviation (s.d.) values. 

\begin{table}[]
\centering
\caption{Summary of settings used in the experiments.}
\begin{tabular}{|l|l|}
\hline 
\textbf{Experimental Parameter} & \textbf{Settings} \\ \hline\hline
\textbf{Training, Testing Split}                                                             & 80\%, 20\%                                                                                         \\ \hline
\textbf{\# of Missing Score Simulations} & 5                                                                                                  \\ \hline
\textbf{\% Missing}                                                                          & \begin{tabular}[c]{@{}l@{}}{[}0, 10, 20, 30, 40,\\ 50, 60, 70, 80, 90{]}\end{tabular}              \\ \hline
\textbf{Univariate Imputations}                                                              & \begin{tabular}[c]{@{}l@{}}Mean\\ Median\end{tabular}                                              \\ \hline
\textbf{Multivariate Imputations}                                                            & \begin{tabular}[c]{@{}l@{}}Bayesian Regression \cite{mackay1992bayesian}\\ Decision Tree \cite{breiman2017classification} \\ K Nearest Neighbors \cite{kramer2013k} \end{tabular} \\ \hline
\textbf{Fusion Applied}                                                                      & Simple Sum Fusion                                                                                  \\ \hline
\end{tabular}
\label{tab:experiments}
\end{table}

\subsection{Experiment 1: Adding a New Modality to an Existing Biometric System}
To simulate adding a new modality to a biometric system, we set all but one modality with complete score data, and incrementally introduce scores from the remaining modality. We consider the scenario where we introduce the best performing modality, or the worst performing modality. For the NIST BSSR1, the best modality for both the identification and verification tasks is the right fingerprint. While Face Algorithm C is the poorest performing modality on the verification task, it is the second best on the identification task.  Face Algorithm G performs relatively poorly on both verification and identification tasks. So we select Face Algorithm G as the worst modality. For the \biocop\ dataset, performance between the 2 modalities are comparable; we consider the Ocular modality the best and the Iris modality the worst performing.

\subsection{Experiment 2: Merging Biometric Systems}
When merging separate biometric systems, it is very unlikely that the systems share the same gallery of subjects. For this reason, the merged score data can be very sparse and scores can be missing randomly across all modalities. For this experiment, there is no constraint on the number of missing modalities other than to ensure that each probe has at least one score. For context, in the NIST BSSR1 with 90\% missing scores, out of the 1,069,156 probes only 145,383 ($\pm$) 257 probes contain scores for all 4 modalities, and up to 296 probes contain only 1 score. 

\subsection{Experiment 3: Retiring a Modality}
The first 2 scenarios describe situations where biometric systems grow larger either through the addition of a new modality or by combining separate biometric systems into one. Conversely, experiment 3 models a scenario where a biometric system loses a modality.  An old sensor might begin malfunctioning or the modality data itself might be outdated. To simulate this situation, we train models on a full training score dataset and use these models to impute missing scores for the missing modalities in the testing set. 

The following describes the questions addressed with this experiment:
\begin{itemize}
\item If a modality is retired, is it better to continue using imputed scores or retrain the biometric system without the retired modality?
\item When a modality is starting to malfunction, is there a limit on the number of times imputation can be applied before it is no longer helpful?
\end{itemize}

\section{Results}
\label{sec:results}

In our experiments, we find that imputation and fusion drastically boosts biometric performance for the NIST BSSR1 dataset, both for the verification and identification tasks. Even when 90\% of probes are incomplete, NIST BSSR1 recognition performance is drastically improved in all three scenarios. The \biocop\ dataset did not show such strong recognition performance gains, however imputation does appear to bridge the gap between a complete score dataset and analysis performed on an incomplete score dataset. We highlight a few of the results below \footnote{Complete results can be viewed \url{https://melissadale.github.io/ICPR2022/}}. 

\subsection{Experiment 1: Adding a New Modality to an Existing Biometric System}

\begin{figure}
\centering
\includegraphics[width=0.4\textwidth]{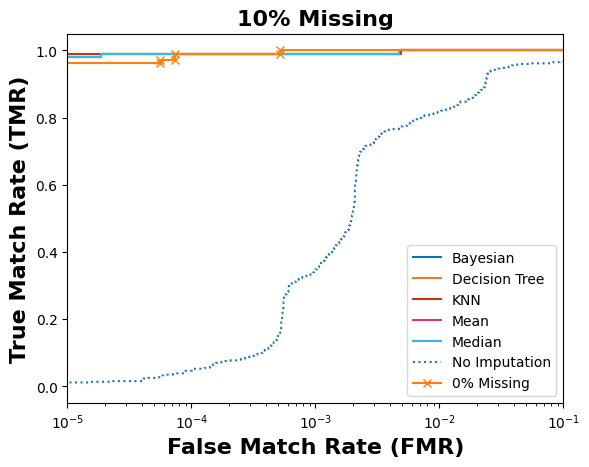}
\caption{ROC curves for imputation methods on the 10\% missing in the NIST BSSR1 dataset scenario in Experiment 1. The curves corresponding to fusion with imputed values overlap near the top with the 0\% missing baseline curve, which is shown in pink.}
\label{fig:ex1-NIST-roc-10}
\end{figure}

\textbf{Verification Tasks} For the NIST BSSR1 dataset, all imputation methods provide recognition performance competitive to the complete dataset, even when 90\% of scores are missing from either the best performing modality (Right Index Fingerprint) or the worst performing modality (Face Comparator G), while the fusion performance on the incomplete dataset is very poor across all amounts of missing data, as can be seen in Figure \ref{fig:ex1-NIST-roc-10}, where only 10\% of the Right Fingerprint modality's scores are missing.  

\begin{figure}
\centering
\includegraphics[width=0.4\textwidth]{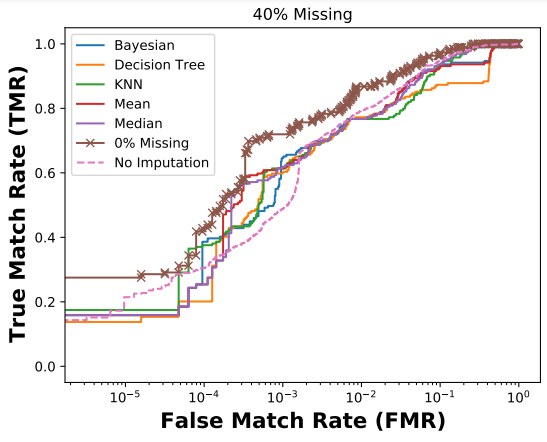}
\caption{ROC curves for imputation methods on the 40\% missing scenario on the \biocop\ dataset in Experiment 1.}
\label{fig:BIO-ex1-40}
\end{figure}

Figure \ref{fig:BIO-ex1-40} provides the ROC plot for the \biocop\ dataset analysis for experiment 1 when 40\% of scores are missing from the ocular modality. We see that no one approach is significantly better than the other, however we see that generally the performance of imputed score values lies between the complete score dataset and the incomplete, non-imputed score dataset. 

These findings suggest that verification performance can be improved when adding a new modality to an existing biometric system, regardless of the individual performance of the modality, by imputing the modality's missing scores before performing fusion.

\begin{figure}
\centering
\includegraphics[width=0.4\textwidth]{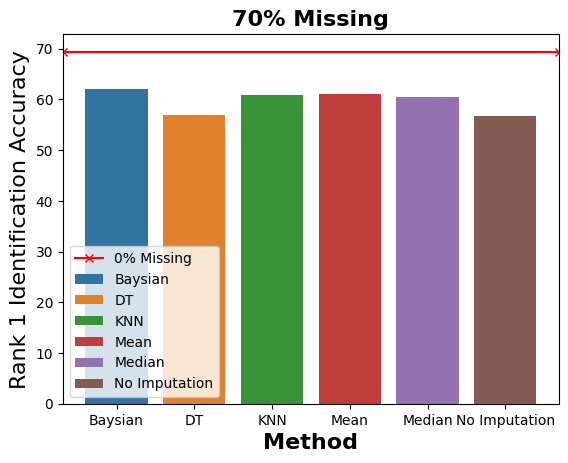}
\caption{Rank 1 accuracies for imputation methods on the 10\% missing scenario for the \biocop\ dataset in Experiment 1.}
\label{fig:ex1-10-r1-bio}
\end{figure}

\textbf{Identification Tasks}
Identification tasks in experiment 1 in particular appear to benefit from analysis with imputed values for both the NIST BSSR1 dataset and the \biocop\ dataset. In fact, the Rank 1 identification accuracy for the \biocop\ dataset is improved over the performance of the complete dataset, as shown in Figure \ref{fig:ex1-10-r1-bio}.

While performance is not maintained across increasing proportions of missing score data, fusion with imputed values consistently outperforms fusion with incomplete score data.

%%%%%%%%%%%%%%%%%%%%%%%%%%%%%%%%%%%%%%%%%%%%%%%%%%%%%%%%%
%%%%%%%%%%%%%%%%%%%%%%%%%%%%%%%%%%%%%%%%%%%%%%%%%%%%%%%%%
%%%%%%%%%%%%%%%%%%%%%%%%%%%%%%%%%%%%%%%%%%%%%%%%%%%%%%%%%

\subsection{Experiment 2: Merging Biometric Systems}
The score data simulated in Experiment 2 turns out to be more complicated than the score data in Experiment 1, because each probe may have up to $N-1$ scores missing, where $N$ is the number of modalities in the biometric system. While the recognition performance is not as strong as the performance in Experiment 1, we can see that both verification and identification tasks benefit from applying imputation methods, even when 90\% of the probe data is incomplete.

\begin{figure}
\centering
\includegraphics[width=0.4\textwidth]{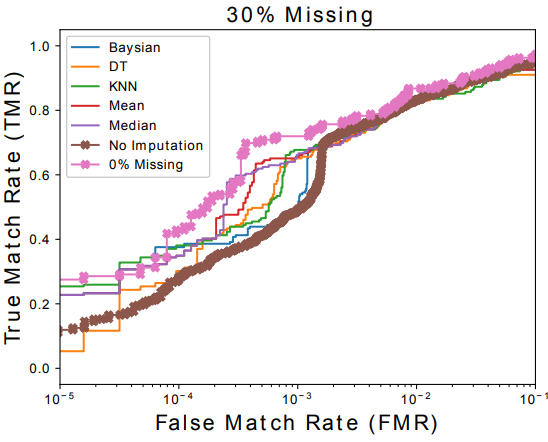}
\caption{ROC curves for imputation methods on the 30\% missing scenario on the \biocop\ in Experiment 2.}
\label{fig:ex2-30-roc-bio}
\end{figure}

\textbf{Verification Tasks}
The verification performance for experiment 2 closely mirrors the verification performance seen in experiment 1 for both NIST BSSR1 dataset and \biocop\ dataset. For example, Figure \ref{fig:ex2-30-roc-bio} shows the ROC plot for the \biocop\ dataset when 30\% of the probes are incomplete.

\begin{figure}
\centering
\includegraphics[width=0.4\textwidth]{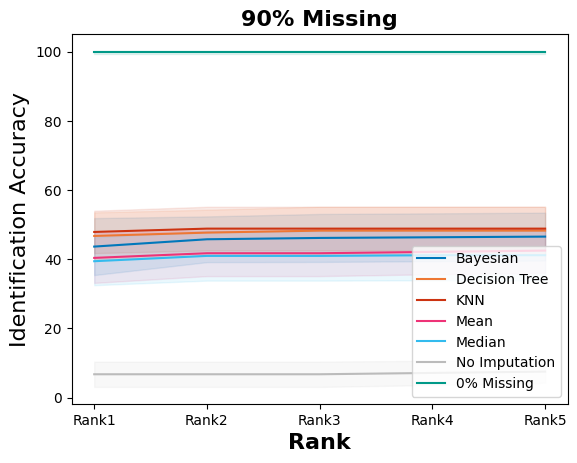}
\caption{Identification accuracies over the first 10 Ranks for the 90\% missing scenario on NIST BSSR1 in Experiment2.}
\label{fig:ex2-90-long}
\end{figure}

\textbf{Identification Tasks}
Figure \ref{fig:ex2-90-long} provides the identification accuracies across the first 10 ranks in the NIST BSSR1 dataset. We can see that when 90\% of probes are incomplete, with some incomplete probes containing only 1 similarity score out of a possible 4 modalities, performance is improved with imputation, especially for the KNN and Decision Tree iterative imputation techniques. 

%%%%%%%%%%%%%%%%%%%%%%%%%%%%%%%%%%%%%%%%%%%%%%%%%%%%%%%%%
%%%%%%%%%%%%%%%%%%%%%%%%%%%%%%%%%%%%%%%%%%%%%%%%%%%%%%%%%
%%%%%%%%%%%%%%%%%%%%%%%%%%%%%%%%%%%%%%%%%%%%%%%%%%%%%%%%%
\subsection{Experiment 3: Retiring a Modality}
\begin{figure}
\centering
\includegraphics[width=0.4\textwidth]{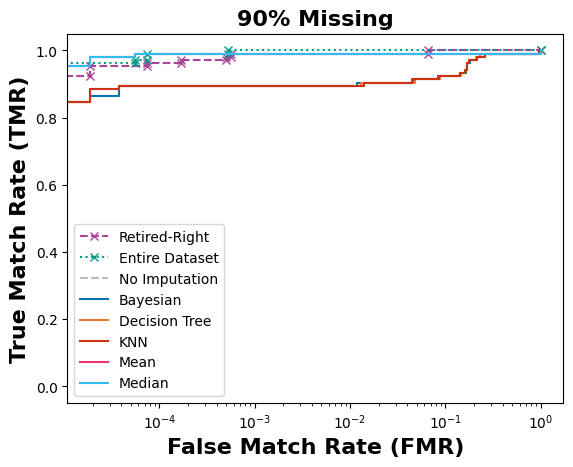}
\caption{ROC plot for NIST BSSR1 dataset for experiment 3 when 90\% of the probes are incomplete.}
\label{fig:ex3-nist-roc}
\end{figure}

When removing a modality from the biometric system, we again see that using imputed scores perform better than applying simple sum fusion without imputation. However, we also see that retraining without the retired modality often yields performances similar to the complete dataset.
% we see that continuing to use an imputed score for this modality instead of dropping the modality altogether from consideration, can continue to provide consistent recognition performance
, as seen in Figure \ref{fig:ex3-nist-roc} for verification tasks and Table \ref{tab:ex3-NIST-90} for identification tasks. This suggest that it may be prudent to retrain a biometric system when retiring a modality.

% These results suggest it may be beneficial to continue imputing a missing modality's \green{similarity} score long after it stops producing \green{similarity} scores. However, after a certain time, it is likely that imputation may cause degradation in performance. 

% \textbf{Verification Tasks}
% Figure \ref{fig:ex3-nist-roc} provides the ROC plots for experiment 3 on the NIST BSSR1 dataset when 10\% of probes contain missing score data. 

% \textbf{Identification Tasks}
% Table \ref{tab:ex3-NIST-90} shows the Rank 1, Rank 2, and Rank3 identification accuracies when 90\% of scores are missing from the probes in the NIST BSSR1 dataset. These results suggest it may be beneficial to continue imputing a missing modality's match score long after it stops producing match scores.

\begin{table}[]
\centering
\caption{The Rank 1, Rank 2, and Rank 3 identification accuracies for the imputation methods on 90\% missing data in Experiment 3 on the NIST BSSR1 dataset.}
\begin{tabular}{|l|c|c|c|}
       \hline
       \textbf{Method} &  \textbf{Rank1}    &  \textbf{Rank2} &  \textbf{Rank3} \\ \hline
    Bayesian &       97.12\%  $\pm$ 1.52 &       98.85\%  $\pm$ 1.05 &       98.85\%  $\pm$ 1.05 \\ \hline 
    Decision Tree &       95.38 \%  $\pm$ 2.49 &       98.85\%  $\pm$ 1.05 &       98.85 \%  $\pm$ 1.05 \\ \hline 
    KNN &       97.12 \%  $\pm$ 1.52 &       98.85 \%  $\pm$ 1.05 &       98.85\%  $\pm$ 1.05 \\ \hline 
    Mean &       98.27\%  $\pm$ 1.43 &       98.85\%  $\pm$ 1.05 &       98.85 \%  $\pm$ 1.05 \\ \hline 
    Median &       98.27 \%  $\pm$ 1.43 &       98.85 \%  $\pm$ 1.05 &       98.85\%  $\pm$ 1.05 \\ \hline
    No Imputation &       11.35\%  $\pm$ 3.15 &       12.12\%  $\pm$ 3.09 &       12.31\%  $\pm$ 3.22 \\\hline
    Retraining  & 99.81        &  99.81       &  99.81       \\\hline
    0\% Missing  & 100       & 100       & 100        \\\hline
\end{tabular}
\label{tab:ex3-NIST-90}
\end{table}

\section{Conclusions}
\label{sec:conclusions}
In this study, we show that applying imputation to incomplete score data in a multimodal biometric system can improve the performance of both verification and identification tasks. The small, well-groomed multimodal NIST BSSR1 dataset benefits greatly from simple sum fusion, with nearly 100\% identification Rank 1 accuracy on the complete dataset. While all imputation methods produced strong recognition performances, iterative imputation with KNN provided the best results in both verification and identification tasks. For the \biocop\ dataset, recognition performance gains over the complete data are not as pronounced, however the recognition performance does not decrease due to the imputed values. Furthermore, in a few situations, fusion with imputed similarity score values outperforms fusion with incomplete score vectors. To better understand how biometric systems are impacted by sparse data, larger multimodal datasets are required. Additionally, multimodal datasets that naturally contain missing score data can better illuminate how these methods perform in situations where the data may not be missing at random.

\section{Future Work}
This study was applied to missing scores, however as mentioned in Section \ref{sec:Background}, it is possible scores are considered missing due poor quality data. Further research is needed to understand the impact of score quality when invoking imputation in multibiometric systems. Additionally, this research has focused on score-level fusion. Further research should be conducted on other levels of the biometric system.

\clearpage
\newpage
\bibliographystyle{./IEEEtran}
\bibliography{./IEEEexample}

\end{document}